\documentclass[letterpaper]{article} 
\usepackage[preprint]{aaai2027}  
\usepackage[hyphens]{url}  
\usepackage{graphicx} 
\usepackage{natbib}  
\usepackage{caption} 
\usepackage{algorithm}
\usepackage{algorithmic}
\usepackage{dsfont}
\usepackage{amsmath}
\usepackage{threeparttable}
\usepackage[table]{xcolor}

\usepackage{newfloat}
\usepackage{listings}
\DeclareCaptionStyle{ruled}{labelfont=normalfont,labelsep=colon,strut=off} 
\floatstyle{ruled}
\newfloat{listing}{tb}{lst}{}
\floatname{listing}{Listing}

\usepackage{booktabs}

\title{CACSurv: Concordance-Aligned Comparative Learning with Large Language Models for Cancer Survival Prediction}
\author{
    Tianqi Xiang,
    Qixiang Zhang,
    Xinpeng Ding,
    Yi Li,
    Xiaomeng Li\corresponding
}
\affiliations{
    The Hong Kong University of Science and Technology\\
    eexmli@ust.hk
}

\begin{document}

\maketitle

\begin{abstract}

Cancer survival prediction supports treatment planning, risk stratification, and follow-up management. Existing methods primarily use structured clinical variables, whole-slide images, genomic profiles, or multimodal combinations, while patient reports remain underexplored as primary inputs. We study report-centric survival prediction using reports that organize pathological, clinical, and molecular evidence into coherent text. Large language models (LLMs) can reason over such reports, but case-wise time regression introduces two fundamental mismatches. First, a formulation mismatch arises because survival evaluation depends on correctly ordering comparable patients, whereas independent time predictions do not enforce inter-patient ranking consistency. Second, a supervision mismatch arises because a censored patient’s observed time only indicates survival beyond that point and cannot serve as an exact regression target, although it still implies orderings relative to patients who died earlier. To address these mismatches, we propose \textbf{CACSurv}, a Concordance-Aligned Comparative framework for report-centric survival prediction. CACSurv reformulates survival modeling as mini-cohort comparative reasoning, where an LLM predicts relative prognostic orderings. We further introduce concordance-aligned rewards derived from comparable relations under right censoring, enabling censored outcomes to provide ranking supervision without exact event-time targets. During inference, Monte Carlo Reference Aggregation compares each evaluation patient with sampled references and aggregates the resulting positions into a cohort-level ranking. We establish \textbf{TCGA-SurvReport}, a benchmark covering six TCGA cancer cohorts. CACSurv achieves the highest C-index on all six cohorts and an average C-index of 0.722, outperforming the strongest published survival model by 6.5 percentage points and the strongest LLM time-regression baseline by 4.2 percentage points. Our code, models, and dataset will be made publicly available at \url{https://github.com/xmed-lab/CACSurv}.
\end{abstract}

\section{Introduction}
\begin{figure*}[t]
    \centering
    \resizebox{1\textwidth}{!}{
    \includegraphics[width=\textwidth]{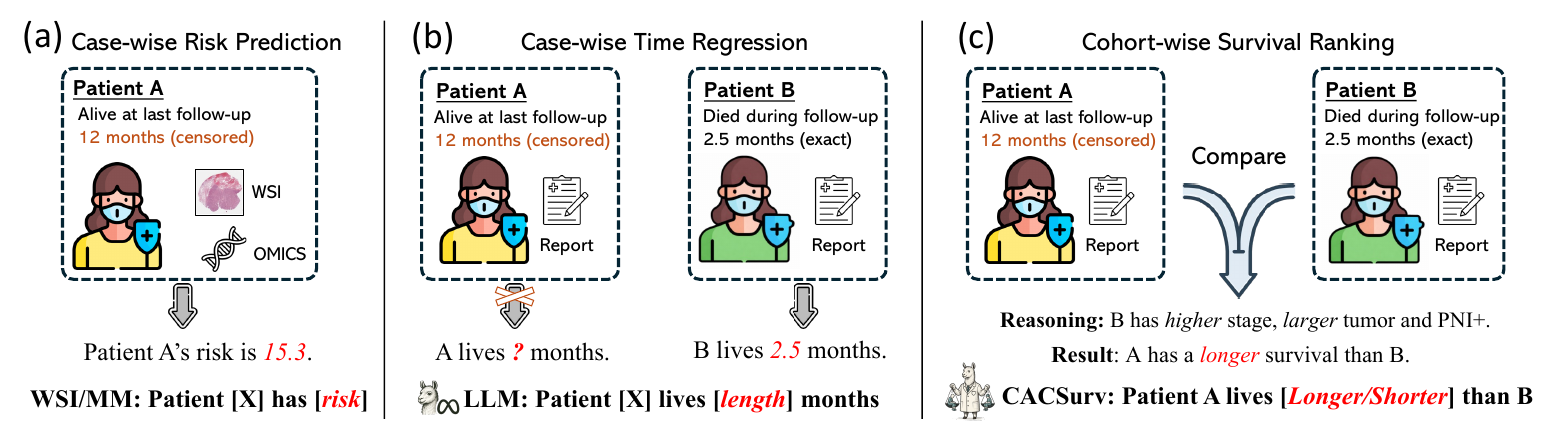}}
    \caption{\textbf{Comparison of survival prediction formulations.} (a) Conventional WSI/multimodal models predict a case-wise risk score. (b) A straightforward report-based LLM performs case-wise time regression, but patients who remain alive at the last follow-up lack exact-time targets. (c) CACSurv directly ranks patients within a mini-cohort with reliable relative comparisons.}
    \label{fig:formulation}
\end{figure*}

Cancer survival prediction aims to characterize a patient’s prognosis from clinical evidence, providing important support for treatment planning, risk stratification, and follow-up management. Existing computational approaches primarily learn from structured clinical variables, whole-slide images (WSIs), genomic profiles, or their multimodal combinations~\cite{transmil,mcat}, typically producing patient-wise risk scores or survival estimates (Fig.~\ref{fig:formulation}(a)). Beyond these commonly studied modalities, patient reports provide a distinct textual interface that organizes clinically interpreted evidence, including diagnosis, staging, pathological findings, and molecular characteristics~\cite{tcgareport}. Despite their information-rich nature, patient reports remain comparatively underexplored as the primary input for cancer survival prediction. In this work, we investigate report-centric cancer survival analysis, where unified patient reports are directly used to derive prognostic predictions.

Recent advances in large language models (LLMs) have enabled effective processing of long-form text and integration of heterogeneous information~\cite{qwen2}, making them a natural candidate for report-centric cancer survival analysis. Recent regression-aware LLM studies have further established pointwise scalar prediction as a viable paradigm for language models~\cite{raft,real}. Following this direction and the patient-wise formulation of conventional survival models (Fig.~\ref{fig:formulation}(a)), a straightforward LLM adaptation is to process each report independently and generate an absolute survival time (Fig.~\ref{fig:formulation}(b)). However, the ability to perform numerical regression does not by itself provide a principled formulation for survival prediction. Although natural for generative models, this case-wise time-regression formulation is not well aligned with either the evaluation or supervision of survival prediction, giving rise to two fundamental mismatches.

First, there is a \textbf{formulation mismatch} between case-wise time regression and concordance-based survival evaluation. As shown in Fig.~\ref{fig:formulation}(b), a time-regression LLM makes separate scalar predictions for Patients A and B. However, survival models are commonly evaluated by the concordance index (C-index), which measures whether comparable patients are ordered correctly rather than the numerical accuracy of individual predictions. Independent time regression lacks explicit constraints on such inter-patient ordering. Moreover, absolute time regression must accommodate substantially different survival time scales across cancer types, as shown in Fig.~\ref{fig:dataset}. Thus, the time-regression formulation focuses on absolute numerical prediction, without directly aligning its objective with the relative prognostic ordering required by survival evaluation.

Second, there is a \textbf{supervision mismatch} between exact-time regression and incomplete survival outcomes. As shown by Patient A in the left example of Fig.~\ref{fig:formulation}(b), being alive at the last follow-up of 12 months indicates only that the patient survived for at least 12 months, while the true survival time remains unknown. Such right-censored cases are common in survival analysis and account for over 60\% of our benchmark (Fig.~\ref{fig:dataset}). Treating the follow-up time as an exact regression target introduces incorrect supervision, while excluding censored patients discards substantial prognostic information. Importantly, these patients still provide valid ordering relations with patients who died before their last follow-up. Thus, censored outcomes are better represented as partial-order supervision than as exact-time labels.

These two mismatches share a common underlying issue: absolute-time regression treats survival as an independent numerical prediction problem, whereas both concordance-based evaluation and the supervision available under right censoring are fundamentally relational. To this end, we propose \textbf{CACSurv}, a Concordance-Aligned Comparative framework for report-centric survival analysis. To handle the formulation mismatch, \textbf{CACSurv introduces a mini-cohort comparative framework that replaces absolute-time prediction with relative prognostic ordering.} The model receives a small group of patient reports and predicts their ordering through comparative reasoning (Fig.~\ref{fig:formulation}(c)). This formulation makes inter-patient ranking the native prediction target and directly aligns model outputs with concordance-based evaluation. During inference, Monte Carlo Reference Aggregation repeatedly compares each evaluation patient with subsets sampled from a shared cancer-specific reference bank and aggregates the resulting relative positions into a reference-relative survival score. These scores place all evaluation patients on a common scale for cohort-level ranking. To resolve the supervision mismatch, \textbf{CACSurv designs concordance-aligned rewards derived from valid comparable relations under right censoring.} Rather than treating the last follow-up time of a censored patient as an exact survival target, these rewards evaluate whether the model correctly predicts the reliable partial-order relations implied by the observed outcomes, allowing censored patients to contribute informative supervision without mislabeling their follow-up times as observed event times or excluding them from training. Together, CACSurv aligns its prediction formulation, supervision signal, and inference procedure with survival concordance. Our main contributions are as follows:
\begin{itemize}
    \item \textbf{Comparative survival framework.} We reformulate report-centric LLM survival analysis from independent absolute-time generation into mini-cohort prognostic ordering, directly aligning model prediction with concordance-based evaluation.
    \item \textbf{Concordance-aligned learning and inference.} We introduce rewards that exploit valid partial-order supervision under right censoring and Monte Carlo Reference Aggregation that converts comparative predictions to cohort-level rankings.
    \item \textbf{Comprehensive benchmark and validation.} We establish TCGA-SurvReport, a six-cohort benchmark that integrates pathological, clinical, and molecular evidence into unified patient reports. CACSurv achieves the highest C-index on all six cohorts, with an average of 0.722, outperforming the strongest published survival model by 6.5 percentage points and the strongest LLM time-regression baseline by 4.2 percentage points.
    
\end{itemize}

\section{Related Work}

\subsection{Unimodal Cancer Survival Analysis}

Single-modality cancer survival analysis has been extensively studied using structured clinical variables and histopathology images. Traditional approaches apply statistical or deep survival models to structured clinical features~\cite{coxph,deepsurv,deephit}. In computational pathology, WSI-based methods commonly aggregate patch-level representations through multiple instance learning~\cite{clam,transmil}, with recent studies further improving slide representation, heterogeneous tissue modeling, and uncertainty estimation~\cite{titan,pamoe,dpsurv}. Despite their architectural differences, these methods generally process each patient independently and produce a patient-specific risk score, hazard estimate, or survival distribution.

\subsection{Multimodal Cancer Survival Analysis}

Multimodal cancer survival analysis integrates histopathology with genomic, transcriptomic, clinical, and textual information to capture complementary prognostic evidence. Early studies model cross-modal interactions through co-attention or optimal-transport-based alignment~\cite{mcat,motcat}, while pathway-aware approaches connect histological representations with biological pathways~\cite{survpath}. More recent methods explore structured cross-modal alignment and latent prognostic representations~\cite{cima,slotspe}. Pathology reports and report-derived embeddings have also been incorporated alongside WSIs, molecular profiles, and clinical variables for multimodal survival prediction~\cite{ps3,songpmlr}.

These multimodal approaches primarily focus on representation learning and information fusion across heterogeneous modalities. When pathology reports are incorporated, they are used as an additional modality alongside WSIs, molecular profiles, and clinical variables. Recent studies have also explored LLM-based prognostic assessment from clinical notes or pathology reports through LLM-assisted extraction of prognostic variables~\cite{kiermeyer2026large}, survival classification at a fixed time point~\cite{phaterpekar2026investigating}, and categorical risk or prognosis prediction~\cite{loeffler2026script,saluja2025cancer}. These studies address related prognostic tasks but differ from censoring-aware survival ranking in their prediction targets and evaluation protocols. CACSurv uses unified patient reports as direct inputs and formulates survival prediction as comparative ranking, replacing independent patient-wise prediction and aligning its supervision and inference with survival concordance under right censoring.

\section{Methodology}
\begin{figure*}[htb]
    \centering
    \resizebox{1\textwidth}{!}{
    \includegraphics[width=\textwidth]{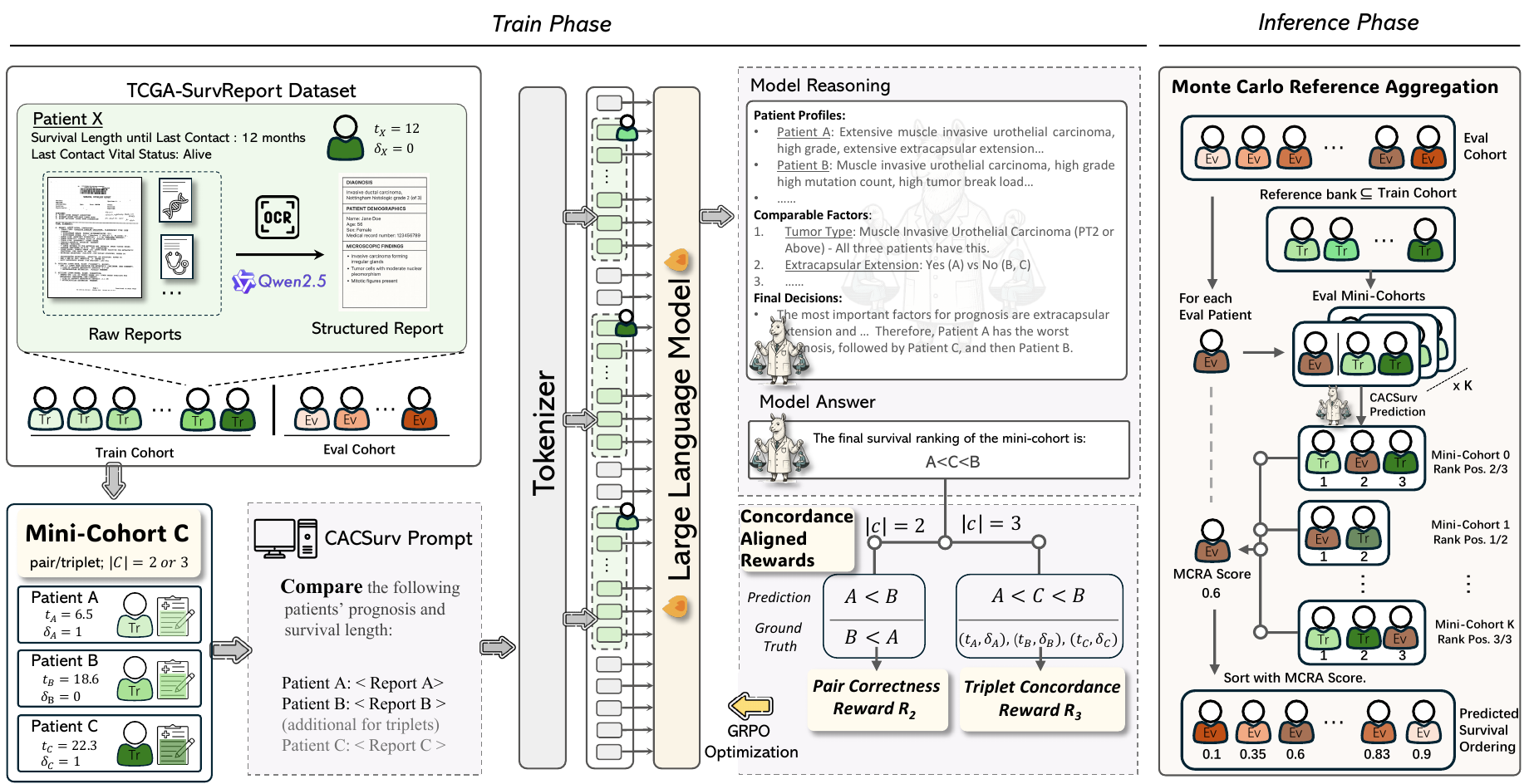}}
    \caption{\textbf{Overview of CACSurv.} During training, CACSurv learns survival orderings over pair and triplet mini-cohorts using concordance-aligned rewards and GRPO. During inference, Monte Carlo Reference Aggregation constructs multiple mini-cohorts for each evaluation patient using randomly sampled reference patients, then averages its normalized rank positions to obtain the final cohort-level survival ordering.}
    \label{fig:main}
\end{figure*}

In this section, we present CACSurv, a concordance-aligned comparative learning framework for report-centric survival prediction, as illustrated in Fig.~\ref{fig:main}. We first formulate the survival prediction problem and briefly review Group Relative Policy Optimization (GRPO)~\cite{deepseekr1}. We then introduce comparative mini-cohort ranking during training and Monte Carlo Reference Aggregation during inference, followed by the pair and triplet concordance-aligned rewards used to optimize CACSurv.

\subsection{Preliminary}\label{method-preliminary}
\noindent\textbf{Problem setup.} 
We consider a cohort of $n$ patients and split it into a training set $\Psi=\{p_1,\dots,p_{n_0}\}$ and an evaluation set $\Psi^{*}=\{p^{*}_1,\dots,p^{*}_{n_1}\}$, where $n_0+n_1=n$.
Each training patient $p_i\in\Psi$ is associated with a patient record $x_i$ and a survival label $(t_i,\delta_i)$, where $t_i$ denotes the observed time and $\delta_i\in\{0,1\}$ indicates whether the event is observed ($\delta_i=1$, patient dead) or the case is right-censored ($\delta_i=0$, patient alive).
For an evaluation patient $p^{*}_j\in\Psi^{*}$, the corresponding record and label are denoted by $(x^{*}_j,t^{*}_j,\delta^{*}_j)$.
Only patient reports are provided to CACSurv as model inputs. The survival outcomes $(t_i,\delta_i)$ of training patients are used solely to construct comparable relations and compute concordance-aligned rewards, and are not included in the model prompt. For evaluation patients, $(t^{*}_j,\delta^{*}_j)$ are used only for performance evaluation after prediction.
Right censoring provides only partial survival information, so we define a patient pair as \emph{comparable} if and only if the patient with the shorter observed time is not censored. We evaluate prognostic performance using the concordance index (C-index), which measures the agreement between predicted preferences and comparable survival outcomes.

\noindent\textbf{Group Relative Policy Optimization (GRPO).}
We employ GRPO~\cite{deepseekr1} to optimize the model with reward-based signals.
Given an input, the current policy $\theta$ samples a set of $G$ candidate outputs $\{o_i\}_{i=1}^{G}$, each associated with a scalar reward $R_i$.
GRPO forms a relative advantage by standardizing each reward within the sampled group:
\begin{equation}
A_i
= \frac{R_i-\operatorname{mean}\big(\{R_j\}_{j=1}^{G}\big)}
{\operatorname{std}\big(\{R_j\}_{j=1}^{G}\big)}.
\label{eq:grpo-adv}
\end{equation}
This group-normalized advantage is shared across all tokens in $o_i$, leading to a clipped surrogate objective:
\begin{equation}
\mathcal{J}_{\mathrm{GRPO}}(\theta)=\frac{1}{G}\sum_{i=1}^{G}\frac{1}{|o_i|}\sum_{t=1}^{|o_i|}
\min\Big(r_{i,t}(\theta)A_{i},\, \bar r_{i,t}(\theta)\,A_{i}\Big),
\label{eq:grpo-loss}
\end{equation}
where $r_{i,t}(\theta)$ is the token-wise likelihood ratio,
$\bar r_{i,t}(\theta)=\operatorname{clip}\big(r_{i,t}(\theta),1-\epsilon,1+\epsilon\big)$, and $|o_i|$ is the length of the $i$-th output.
We omit the KL term here for brevity.

\subsection{CACSurv for Comparative Survival Ranking}\label{method-llarm}
CACSurv reformulates report-centric survival prediction as a comparison-based ranking problem. Instead of treating each patient as an atomic unit and predicting an absolute survival time or continuous risk score, CACSurv uses mini-cohorts as the basic prediction units and directly learns relative survival orderings among patients.

\noindent\textbf{Mini-cohort construction.}
A mini-cohort is a small set of patients sampled from the full cohort, used as the basic prediction unit in CACSurv. For training, we sample an index set $I_k \subseteq \{1,\dots,n_0\}$ for the $k$-th mini-cohort $C_k$ from the training set $\Psi$:
\begin{equation}
C_k=\{p_i : i\in I_k\},\quad m_k=|I_k|.
\label{eq:minicohort-def-new}
\end{equation}
Each mini-cohort consists of patients from the same cancer type, with no duplicate patients within the cohort.

For evaluation, we fix a cancer-type-specific reference bank $\psi \subseteq \Psi$ containing training patients of the same cancer type as the evaluation patient. For each evaluation patient $p^{*}_j\in\Psi^{*}$, we construct $K$ evaluation mini-cohorts by combining $p^{*}_j$ with a subset of reference patients from $\psi$:
\begin{equation}
C^{*}_{j,l}=\{p^{*}_j\}\cup R_{j,l},\quad R_{j,l}\subseteq \psi,\quad l=1,\dots,K.
\label{eq:eval-minicohort}
\end{equation}
Given any ordering $y$ over a mini-cohort $C$, we write $u \succ_{y} v$ if $u$ is preferred to $v$ under $y$, meaning $u$ is predicted to have a longer survival than $v$.

\noindent\textbf{Training phase.}
Given a training mini-cohort $C_k$, let $X_k = \{x_i : p_i \in C_k\}$ denote its corresponding patient reports. CACSurv predicts a survival ordering
\begin{equation}
\hat{y}_k = f_\theta(X_k),
\label{eq:model-output-new}
\end{equation}
where $\hat{y}_k$ orders the patients in $C_k$ from shorter to longer predicted survival.
We optimize $f_\theta$ with GRPO using the concordance-aligned rewards defined below.
Repeated training over different mini-cohorts exposes the model to inter-patient survival relations and directly encourages concordance-consistent ranking behavior.

\noindent\textbf{Inference phase (Monte Carlo Reference Aggregation).}
For each evaluation mini-cohort $C^{*}_{j,l}$, CACSurv takes the corresponding patient reports $X^*_{j,l}$ as input and predicts an ordering $\hat{y}^{*}_{j,l} = f_\theta(X^{*}_{j,l})$. We compute the normalized rank position of the evaluation patient $p^{*}_j$ as:
\begin{equation}
s(p^{*}_j; C^{*}_{j,l}, \hat{y}^{*}_{j,l})
= \frac{\left| \left\{ u \in C^{*}_{j,l}\setminus\{p^{*}_j\} \;:\; p^{*}_j \succ_{\hat{y}^{*}_{j,l}} u \right\} \right|}{|C^{*}_{j,l}|-1}.
\label{eq:local-position}
\end{equation}
We aggregate its relative positions across $K$ mini-cohorts:
\begin{equation}
S(p^{*}_j)=\frac{1}{K}\sum_{l=1}^{K} s(p^{*}_j; C^{*}_{j,l}, \hat{y}^{*}_{j,l}).
\label{eq:global-position}
\end{equation}
A larger $S(p^{*}_j)$ indicates longer predicted survival. The final full cohort-level ordering is obtained by sorting the evaluation patients in ascending order of $S(p^{*}_j)$. 

Under uniform random sampling from the reference bank, each normalized rank position is a bounded Monte Carlo observation of the patient's reference-relative position. Therefore,
\begin{equation}
E[S(p^{*}_j)] = \mu_j = E_{R\sim\psi}[s(p_j^*;C_R, \hat{y}_R)],
\end{equation}
and $S(p^{*}_j)$ converges to $\mu_j$ as $K$ increases. When the predicted preferences are consistent across mini-cohorts, $\mu_j$ corresponds to the proportion of reference patients predicted to have shorter survival than $p^*_j$, providing a reference-relative approximation of its global position.

MCRA avoids direct comparisons among evaluation patients and assigns each patient a fixed number of comparisons independent of the evaluation cohort size, allowing it to support both small evaluation cohorts and sequential patient arrival.

\subsection{Concordance-aligned Rewards}\label{method-reward}
Our concordance-aligned rewards convert censored survival outcomes into reliable preference supervision. If a patient is right-censored at time $t_i$, the patient is known to have survival longer than any patient who died before $t_i$, even though the exact survival time remains unknown. Based on such comparable relations, we define a pair correctness reward for pair mini-cohorts and a triplet concordance reward for mini-cohorts of size three or beyond.

\noindent\textbf{Pair correctness reward ($|C|=2$).}
For a pair mini-cohort $C=\{p_i,p_j\}$, we ensure during construction that the two patients are comparable. Given a predicted ordering $\hat{y}$, we define the correctness reward as
\begin{small}
\begin{equation}
R_2(C,\hat{y})
=
\mathds{1}\!\big[t_i<t_j\big]\cdot \mathds{1}\!\big[p_j \succ_{\hat{y}} p_i\big]
+
\mathds{1}\!\big[t_j<t_i\big]\cdot \mathds{1}\!\big[p_i \succ_{\hat{y}} p_j\big].
\label{eq:pair-reward-new}
\end{equation}
\end{small}
Since pair comparability is enforced during construction, the event indicator is omitted from Eq.~\ref{eq:pair-reward-new}.

\noindent\textbf{Triplet concordance reward ($|C| \ge 3$).}
For a triplet mini-cohort $C$, right censoring may make only a subset of patient pairs comparable.
We define the set of comparable directed relations within $C$ as
\begin{equation}
\mathcal{V}(C)
=
\left\{(p_u,p_v)\in C\times C \;:\; u\neq v,\; t_v<t_u,\; \delta_v=1 \right\}.
\label{eq:valid-pairs-new}
\end{equation}
The triplet concordance reward encourages ordering consistency over all comparable relations in $\mathcal{V}(C)$:
\begin{small}
\begin{equation}
R_3(C,\hat{y})
=
\begin{cases}
\frac{1}{|\mathcal{V}(C)|}\sum\limits_{(p_u,p_v)\in \mathcal{V}(C)} \mathds{1}[\,p_u \succ_{\hat{y}} p_v\,], & |\mathcal{V}(C)|>0,\\[6pt]
0, & |\mathcal{V}(C)|=0.
\end{cases}
\label{eq:triplet-reward-new}
\end{equation}
\end{small}
Note that this formulation also naturally accepts mini-cohorts with more than three patients by defining $\mathcal{V}(C)$ over all within-cohort comparable relations.

\noindent Therefore, we define the unified concordance-aligned reward as
\begin{equation}
R(C,\hat{y})
=
\mathds{1}[|C|=2]\cdot R_2(C,\hat{y})
+
\mathds{1}[|C|\ge 3]\cdot R_3(C,\hat{y}),
\label{eq:mix-reward-new}
\end{equation}
which is used as the reward signal for GRPO optimization during training.

\section{Experiments}
\begin{table*}[tb]
  \centering
  
  \setlength{\tabcolsep}{3.5pt}
  \renewcommand{\arraystretch}{1.0}
  \resizebox{\textwidth}{!}{ %
  \begin{tabular}{l|cccccc|c}
    \toprule
    \textbf{Method} & \textbf{BLCA} & \textbf{BRCA} & \textbf{COADREAD} & \textbf{HNSC} & \textbf{LUAD} & \textbf{STAD} & \textbf{AVG@6} \\
    \midrule

    \multicolumn{8}{c}{\textbf{\textit{WSI/Multimodal Survival Models}}} \\
    \midrule
    TransMIL & $0.571_{\pm 0.083}$ & $0.640_{\pm 0.056}$ & $0.642_{\pm 0.084}$ & $0.568_{\pm 0.068}$ & $0.548_{\pm 0.027}$ & $0.599_{\pm 0.052}$ & $0.595_{\pm 0.039}$ \\
    TITAN  & $0.625_{\pm 0.065}$ & $0.666_{\pm 0.033}$ & $0.649_{\pm 0.060}$ & $0.586_{\pm 0.033}$ & $0.607_{\pm 0.048}$ & $0.548_{\pm 0.074}$ & $0.613_{\pm 0.043}$ \\
    PAMoE  & $0.646_{\pm 0.054}$ & $0.704_{\pm 0.038}$ & $0.702_{\pm 0.050}$ & $0.620_{\pm 0.032}$ & $0.646_{\pm 0.035}$ & $0.612_{\pm 0.070}$ & $0.655_{\pm 0.036}$ \\
    DPSurv  & $0.611_{\pm 0.064}$ & $0.600_{\pm 0.055}$ & $0.585_{\pm 0.089}$ & $0.531_{\pm 0.074}$ & $0.570_{\pm 0.020}$ & $0.536_{\pm 0.041}$ & $0.572_{\pm 0.033}$ \\
    MCAT  & $0.620_{\pm 0.012}$ & $0.564_{\pm 0.026}$ & $0.612_{\pm 0.052}$ & $0.518_{\pm 0.024}$ & $0.571_{\pm 0.031}$ & $0.528_{\pm 0.075}$ & $0.569_{\pm 0.042}$ \\
    MOTCat  & $0.623_{\pm 0.012}$ & $0.570_{\pm 0.041}$ & $0.617_{\pm 0.030}$ & $0.548_{\pm 0.018}$ & $0.561_{\pm 0.022}$ & $0.531_{\pm 0.075}$ & $0.575_{\pm 0.050}$ \\
    SurvPath & $0.581_{\pm 0.038}$ & $0.623_{\pm 0.086}$ & $0.657_{\pm 0.095}$ & $0.539_{\pm 0.032}$ & $0.571_{\pm 0.035}$ & $0.562_{\pm 0.075}$ & $0.589_{\pm 0.043}$ \\
    PS3 & $0.638_{\pm 0.075}$ & $0.632_{\pm 0.056}$ & $0.613_{\pm 0.083}$ & $0.622_{\pm 0.023}$ & $0.658_{\pm 0.033}$ & $0.556_{\pm 0.049}$ & $0.620_{\pm 0.061}$ \\
    CIMA  & $0.605_{\pm 0.082}$ & $0.607_{\pm 0.010}$ & $0.643_{\pm 0.054}$ & $0.576_{\pm 0.027}$ & $0.609_{\pm 0.034}$ & $0.542_{\pm 0.092}$ & $0.597_{\pm 0.034}$ \\
    SlotSPE & $0.658_{\pm 0.007}$ & $0.709_{\pm 0.009}$ & $0.710_{\pm 0.006}$ & $0.607_{\pm 0.016}$ & $0.656_{\pm 0.016}$ & $0.601_{\pm 0.009}$ & $0.657_{\pm 0.045}$ \\
    \midrule

    \multicolumn{8}{c}{\textbf{\textit{Report-centric Survival Models}}} \\
    \midrule
    Tabular-CoxPH    & $0.623_{\pm 0.046}$ & $0.619_{\pm 0.071}$ & $0.683_{\pm 0.045}$ & $0.591_{\pm 0.035}$ & $0.654_{\pm 0.048}$ & $0.598_{\pm 0.085}$ & $0.628_{\pm 0.032}$ \\
    Tabular-DeepHit  & $0.618_{\pm 0.034}$ & $0.651_{\pm 0.046}$ & $0.696_{\pm 0.041}$ & $0.589_{\pm 0.034}$ & $0.671_{\pm 0.034}$ & $0.621_{\pm 0.078}$ & $0.641_{\pm 0.036}$ \\
    PubMedBERT-DeepSurv & $0.493_{\pm 0.092}$ & $0.609_{\pm 0.062}$ & $0.596_{\pm 0.051}$ & $0.564_{\pm 0.072}$ & $0.595_{\pm 0.025}$ & $0.507_{\pm 0.029}$ & $0.561_{\pm 0.045}$ \\
    BioMistral-CoxPH  & $0.489_{\pm 0.046}$ & $0.582_{\pm 0.052}$ & $0.615_{\pm 0.037}$ & $0.515_{\pm 0.047}$ & $0.528_{\pm 0.061}$ & $0.539_{\pm 0.066}$ & $0.545_{\pm 0.046}$ \\
    Qwen2.5-7B Time-ZS & $0.504_{\pm 0.009}$ & $0.693_{\pm 0.036}$ & $0.669_{\pm 0.024}$ & $0.518_{\pm 0.010}$ & $0.573_{\pm 0.035}$ & $0.553_{\pm 0.029}$ & $0.585_{\pm 0.072}$ \\
    Qwen2.5-72B Time-ZS & $0.646_{\pm 0.053}$ & $0.711_{\pm 0.051}$ & $0.745_{\pm 0.043}$ & ${0.646_{\pm 0.025}}$ & $0.663_{\pm 0.060}$ & ${0.670_{\pm 0.050}}$ & $0.680_{\pm 0.047}$ \\
    Qwen2.5-7B Time-SFT  & $0.629_{\pm 0.049}$ & $0.607_{\pm 0.093}$ & $0.649_{\pm 0.064}$ & $0.620_{\pm 0.048}$ & $0.615_{\pm 0.020}$ & $0.634_{\pm 0.038}$ & $0.626_{\pm 0.013}$ \\

    \midrule
    CACSurv-7B (Pair-only) & $\underline{0.661_{\pm 0.060}}$ & $\underline{0.745_{\pm 0.074}}$ & $\underline{0.766_{\pm 0.057}}$ & $\underline{0.667_{\pm 0.040}}$ & $\underline{0.687_{\pm 0.046}}$ & $\underline{0.683_{\pm 0.054}}$ & $\underline{0.702_{\pm 0.040}}$ \\
    \textbf{CACSurv-7B} & $\mathbf{0.685_{\pm 0.061}}$ & $\mathbf{0.763_{\pm 0.063}}$ & $\mathbf{0.806_{\pm 0.058}}$ & $\mathbf{0.689_{\pm 0.046}}$ & $\mathbf{0.693_{\pm 0.064}}$ & $\mathbf{0.698_{\pm 0.047}}$ & $\mathbf{0.722_{\pm 0.046}}$ \\
    \bottomrule
  \end{tabular}%
  }
  
   \caption{Main results from 5-fold cross-validation on TCGA-SurvReport (C-index, mean$\pm$std). Time-ZS and Time-SFT denote direct survival-time prediction using zero-shot inference and supervised fine-tuning, respectively. Pair-only uses only pair mini-cohorts during both training and inference. Best results are shown in \textbf{bold}, and second-best results are \underline{underlined}.}
  \label{tab:main}
\end{table*}

\subsection{Dataset Construction and Cohort Statistics}
We construct TCGA-SurvReport, a report-centric survival dataset covering six TCGA cancer cohorts: BLCA, BRCA, COADREAD, HNSC, LUAD, and STAD. For each patient, we collect OCR-derived pathology text from TCGA-Reports~\cite{tcgareport}, clinical records, and molecular profiles, which provide complementary pathological, demographic, clinical, and molecular information. The three sources are aligned using TCGA patient identifiers.

The released OCR text follows heterogeneous report formats and contains duplicated content, administrative statements, and detailed specimen-processing descriptions. To address these issues, we use Qwen2.5-72B-Instruct with a fixed extractive prompt to remove irrelevant content and organize the retained information into a standardized patient report. The model is prohibited from adding diagnoses, medical facts, prognostic interpretations, or clinical implications absent from the source records. We manually reviewed a sampled subset of processed reports to assess their factual consistency with the corresponding source records.

\begin{figure}[htb]
    \centering
    \includegraphics[width=1\columnwidth]{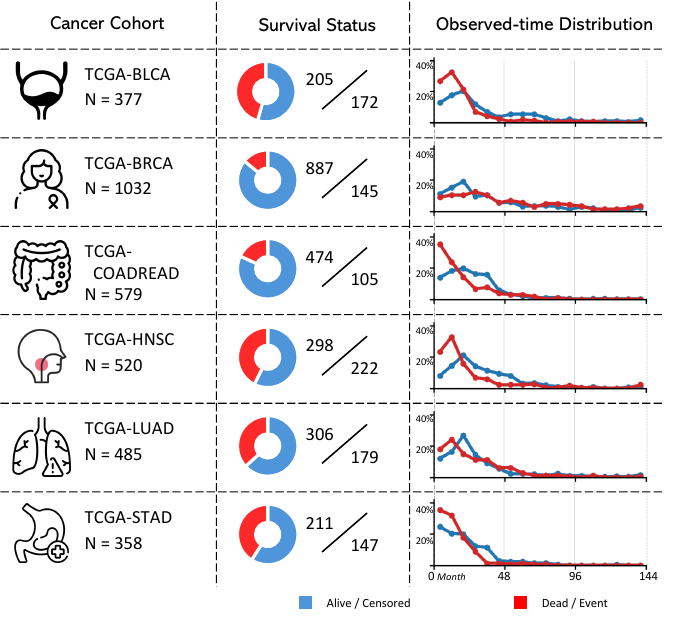}
    \caption{\textbf{Cohort statistics of TCGA-SurvReport.} For each cancer cohort, we report the number of patients, the proportions of alive/censored and dead/event cases, and their observed-time distributions. Observed time denotes the last follow-up time for censored patients and the event time for deceased patients. Each curve shows the percentage of patients within the corresponding status group in each time interval.}
    \label{fig:dataset}
\end{figure}

Observed time and vital status are extracted separately as survival labels and are not included in the unified patient reports. We retain patients with both a valid unified report and an available survival record. Fig.~\ref{fig:dataset} summarizes the cohort statistics across the six cancer types.

\subsection{Baselines}
We compare CACSurv with representative WSI-based and multimodal survival models, including TransMIL~\cite{transmil}, TITAN~\cite{titan}, MCAT~\cite{mcat}, MOTCat~\cite{motcat}, SurvPath~\cite{survpath}, PS3~\cite{ps3}, PAMoE~\cite{pamoe}, CIMA~\cite{cima}, SlotSPE~\cite{slotspe}, and DPSurv~\cite{dpsurv}. These baselines cover WSI-only, histology–molecular, and histology–report–molecular survival prediction settings. For report-centric comparison, we include Tabular-CoxPH~\cite{coxph}, Tabular-DeepHit~\cite{deephit}, PubMedBERT-DeepSurv~\cite{deepsurv}, and BioMistral-CoxPH~\cite{songpmlr}. We further construct LLM-based time-regression baselines using zero-shot inference with Qwen2.5-7B/72B-Instruct~\cite{qwen2} and supervised fine-tuning with Qwen2.5-7B-Instruct.

\subsection{Implementation and Evaluation Protocol}
All methods are evaluated using the same patient cohorts and identical patient-level five-fold cross-validation splits. Models are trained only on the training patients of each fold, and performance is reported as the mean and standard deviation of the C-index across the five folds. All baseline results are obtained by rerunning the corresponding methods under this unified evaluation protocol.

For WSI-based methods, we use CONCHv1.5~\cite{conch} to extract patch features. PS3 uses PLIP~\cite{plip} as it requires aligned image and text encoders. Molecular inputs are processed following the original implementation of each method. Tabular-CoxPH and Tabular-DeepHit use structured prognostic variables extracted from the reports, while PubMedBERT-DeepSurv and BioMistral-CoxPH use report embeddings produced by PubMedBERT~\cite{pubmedbert} and BioMistral~\cite{biomistral}, respectively.

CACSurv uses Qwen2.5-7B-Instruct as the backbone. For each fold, we sample 1,000 pair and 1,000 triplet mini-cohorts from each cancer type and optimize the model for one epoch. Each pair is constructed to be comparable, and each triplet contains at least one comparable patient pair. During training, observed time and vital status are used only to enforce these sampling constraints and compute the concordance-aligned rewards; only patient reports are provided to the model. During inference, the training patients form the reference bank, and each test patient is compared against $K = 40$ randomly sampled reference mini-cohorts. Test survival outcomes, including observed time and vital status, are used only after prediction to calculate the final C-index.

\subsection{Comparisons with State-of-the-Art Methods}
Table~\ref{tab:main} presents the survival prediction results across the six cancer cohorts. CACSurv achieves the highest C-index on every cohort and the best average C-index of 0.722. Its simplified pair-only variant consistently ranks second, with an average C-index of 0.702.

Among existing published survival models, SlotSPE achieves the strongest average performance of 0.657. CACSurv exceeds SlotSPE by 6.5 percentage points and consistently outperforms the evaluated WSI-based, multimodal, and conventional report-centric survival models. These results highlight the strong prognostic value of unified patient reports and the effectiveness of comparative survival modeling.

We further construct LLM-based time-regression baselines that directly predict absolute survival time from each report. CACSurv with a 7B backbone outperforms the strongest of these baselines, Qwen2.5-72B Time-ZS, by 4.2 percentage points, indicating that comparative prediction is more effective for this task than direct absolute-time regression.

\begin{table}[htb]
    \centering
    \setlength{\tabcolsep}{14pt}
    \begin{tabular}{l|cc}
        \toprule
        \textbf{Setting} & Eval.Pair & \textbf{Eval.Mix} \\
        \midrule
        Train.Pair & 0.702 & 0.703 \\
        \textbf{Train.Mix} & 0.715 & \textbf{0.722} \\
        \bottomrule
    \end{tabular}
    \caption{Ablation on pair and triplet mini-cohort instantiations during training and inference. Mix means Pair+Triplet.
    }
    \label{tab:instantiation}
\end{table}

\begin{table}[htb]
    \centering
    \setlength{\tabcolsep}{2.5pt}
    \begin{tabular}{l|cccc}
        \toprule
        \textbf{Optimization} &
        \textbf{BLCA} &
        \textbf{COADREAD} &
        \textbf{HNSC} &
        \textbf{AVG@6} \\
        \midrule
        Rank-SFT & 0.656 & 0.740 & 0.653 & 0.682 \\
        \textbf{Ours} & \textbf{0.685} & \textbf{0.806} & \textbf{0.689} & \textbf{0.722} \\
        \bottomrule
    \end{tabular}
    \caption{Ablation of ranking-based SFT and our concordance -aligned reward optimization under the same comparative formulation.}
    \label{tab:reward_vs_sft}
\end{table}

\subsection{Ablation on Pair and Mixed Mini-Cohorts}
We evaluate pair and triplet mini-cohort compositions during both training and inference in Table~\ref{tab:instantiation}. Train.Pair trains with 2,000 pair mini-cohorts, whereas Train.Mix trains with 1,000 pairs and 1,000 triplets, keeping the total number of training mini-cohorts, backbone, data splits, and training epochs unchanged. Pair-and-triplet training consistently outperforms pair-only training, improving the average C-index from 0.702 to 0.715 under pair inference and from 0.703 to 0.722 under pair-and-triplet inference. The best performance is achieved when mixed mini-cohorts are used in both stages, indicating that triplet relations provide complementary comparative supervision and can be further exploited during inference.

\subsection{Ablation on Reward-based Optimization}
We compare Rank-SFT and concordance-aligned reward optimization under identical settings, varying only the optimization method. Rank-SFT directly fine-tunes the model to generate the target ranking, whereas CACSurv optimizes the concordance-aligned rewards with GRPO. Table~\ref{tab:reward_vs_sft} reports three representative cohorts and the six-cohort average for compactness, while CACSurv consistently outperforms Rank-SFT across all six cohorts. Overall, concordance-aligned reward optimization improves the average C-index from 0.682 to 0.722, demonstrating its effectiveness beyond directly supervising the target ranking.

\begin{table}[htb]
    \centering
    \setlength{\tabcolsep}{3.5pt}
    \resizebox{\columnwidth}{!}{%
        \begin{tabular}{l|cccc}
            \toprule
            \textbf{Setting} &
            \textbf{$K=10$} &
            \textbf{$K=20$} &
            \textbf{$K=40$} &
            \textbf{$K=80$} \\
            \midrule
            Train.Pair, Eval.Pair
            & 0.692 & 0.700 & 0.702 & 0.704 \\

            Train.Mix, Eval.Pair
            & 0.701 & 0.707 & 0.715 & 0.718 \\

            \textbf{Train.Mix, Eval.Mix}
            & \textbf{0.706}
            & \textbf{0.715}
            & \textbf{0.722}
            & \textbf{0.723} \\
            \bottomrule
        \end{tabular}%
    }
    \caption{Sensitivity to the number of reference comparisons $K$ during inference. Mix means Pair+Triplet.}
    \label{tab:k_sensitivity}
\end{table}

\subsection{Sensitivity to Number of Reference Comparisons}
Table~\ref{tab:k_sensitivity} reports the sensitivity to the number of reference comparisons $K$. We keep the trained models and all other inference settings unchanged for fairness. Performance improves consistently as $K$ increases and largely converges at $K=40$. Under CACSurv's default setting (Train.Mix and Eval.Mix), further increasing K from 40 to 80 only improves the C-index from 0.722 to 0.723. We therefore adopt $K=40$ as an acceptable accuracy-complexity tradeoff.

\begin{figure}[htb]
    \centering
    \includegraphics[width=0.95\columnwidth]{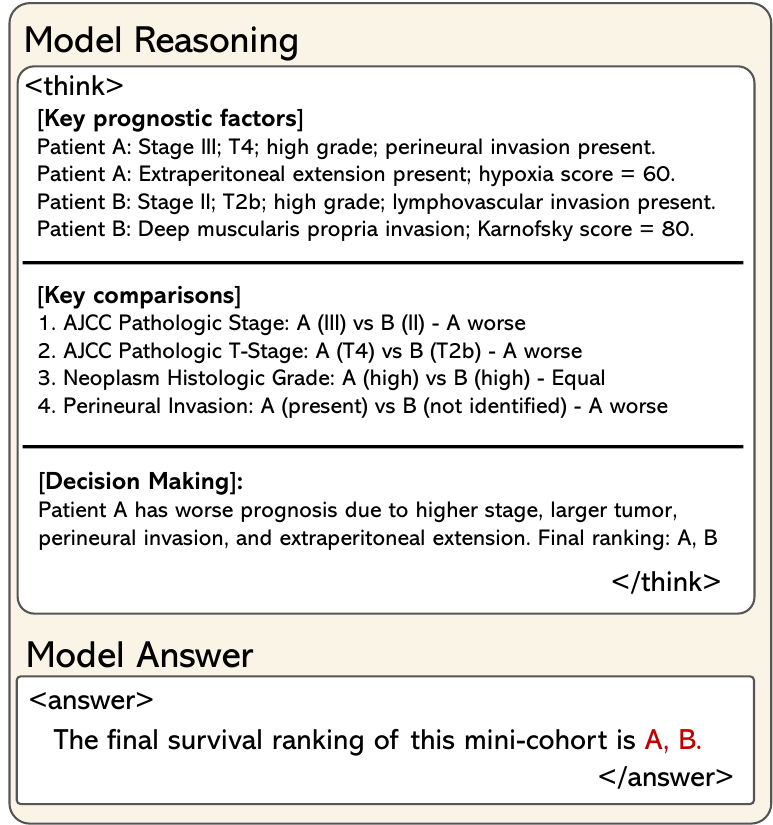}
    \caption{\textbf{Example CACSurv rollout on a pair mini-cohort.} In the \texttt{<think>} section, the model identifies patient-specific prognostic factors, compares discriminative evidence across patients, and derives a ranking decision. The predicted survival ordering is then returned in the \texttt{<answer>} section.}
    \label{fig:rollout}
\end{figure}

\subsection{CACSurv Rollout Example}
Fig.~\ref{fig:rollout} presents a CACSurv rollout example for a pair mini-cohort. The model first identifies patient-specific prognostic factors, then compares the two patients based on the extracted evidence, and finally produces the survival ranking. This example shows how CACSurv converts report evidence into an explicit comparative prediction.

\section{Conclusion}
We present CACSurv, a concordance-aligned comparative framework for report-centric cancer survival prediction. CACSurv addresses the formulation and supervision mismatches of conventional LLM-based time regression by directly predicting prognostic orderings within mini-cohorts and optimizing concordance-aligned rewards derived from valid comparable relations under right censoring. Monte Carlo Reference Aggregation further converts these local comparative predictions into stable cohort-level survival rankings. We also establish TCGA-SurvReport, a unified report-centric benchmark spanning six TCGA cancer cohorts. Under identical patient-level five-fold cross-validation splits, CACSurv achieves the best performance among all evaluated methods on every cohort, with an average C-index of 0.722, exceeding the strongest published survival model by 6.5 percentage points and the strongest LLM time-regression baseline by 4.2 percentage points. Further analyses show that pair and triplet mini-cohorts provide complementary benefits during training and inference, and concordance-aligned reward optimization consistently outperforms ranking-based SFT. These findings demonstrate that aligning the prediction target, learning signal, and inference procedure with survival concordance provides an effective approach to LLM-based survival analysis under right censoring.

\bibliography{reference}

\end{document}